\documentclass[runningheads]{llncs}
\usepackage{graphicx}
\usepackage{amsmath}
\usepackage{graphicx}
\usepackage{multirow}
\usepackage[colorlinks=true, allcolors=blue]{hyperref}
\usepackage{tikzsymbols}
\usepackage{subfig}
\usepackage{stfloats}
\usepackage{boldline}
\usepackage{comment}
\usepackage{cite}
\usepackage{color}

\definecolor{myred}{rgb}{.8,.0,.0}

\newcommand{\aucall}[1]{0.83}
\newcommand{\aucmin}[1]{0.76}
\newcommand{\aucmax}[1]{0.91}

\usepackage[inline]{enumitem}
\usepackage{graphicx}
\usepackage{caption}

\begin{document}
\title{Risk of Training Diagnostic Algorithms on Data with Demographic Bias}
\titlerunning{Bias in Diagnostic Algorithms}


%
\author{Samaneh Abbasi-Sureshjani$^{1}$, Ralf Raumanns$^{1,2}$, Britt E. J. Michels$^{1}$,\\ Gerard Schouten$^{1,2}$ and Veronika Cheplygina$^{1}$}

\institute{Eindhoven University of Technology, The Netherlands \\ \email{\{s.abbasi, r.raumanns, v.cheplygina\}@tue.nl} \and 
Fontys University of Applied Science, The Netherlands \\
\email{g.schouten@fontys.nl}
}

\authorrunning{S. Abbasi-Sureshjani et al.}



\maketitle

\begin{abstract}
One of the critical challenges in machine learning applications is to have fair predictions. There are numerous recent examples in various domains that convincingly show that algorithms trained with biased datasets can easily lead to erroneous or discriminatory conclusions.  This is even more crucial in clinical applications where predictive algorithms are designed mainly based on a given set of medical images, and demographic variables such as age, sex and race are not taken into account. In this work, we conduct a survey of the MICCAI 2018 proceedings to investigate the common practice in medical image analysis applications. Surprisingly, we found that papers focusing on diagnosis rarely describe the demographics of the datasets used, and the diagnosis is purely based on images. In order to highlight the importance of considering the demographics in diagnosis tasks, we used a publicly available dataset of skin lesions. We then demonstrate that a classifier with an overall area under the curve (AUC) of \aucall, has variable performance between \aucmin~ and \aucmax~ on subgroups based on age and sex, even though the training set was relatively balanced. Moreover, we show that it is possible to learn unbiased features by explicitly using demographic variables in an adversarial training setup, which leads to balanced scores per subgroups. 
Finally, we discuss the implications of these results and provide recommendations for further research.  

\keywords{Computer-aided diagnosis \and Demographic bias \and Classification parity}

\end{abstract}

\section{Introduction}
In medical image analysis, machine learning algorithms can be on par with or even exceed the performance of experts.  However, for reliable generalization, large datasets are needed, that are representative of the population on which they are ultimately applied. In the medical domain, this is often not the case ~\cite{cheplygina2019not,greenspan2016guest}. A further requirement is that the properties of the training data are similar to the test data, which is sometimes overlooked. For example, some patient groups (based on age, sex, ethnicity among others) can be overrepresented in the data, biasing the model. Besides the notorious discriminatory face recognition example \cite{buolamwini2018gender}, detrimental effects of such bias have been demonstrated in various domains, varying from predictions of recidivism, to job offers or loan decisions. For medical imaging, the problem seems relatively unexplored, despite the potentially harmful consequences. 

We aim to quantify whether and how bias is addressed in medical imaging papers focusing on the diagnosis. We first survey proceedings from a recent conference. For selected papers, we report the sample size, whether any demographic measures are available, whether these are used by the algorithm and whether demographics/bias are discussed in the paper. Using a dataset of skin lesions, we then demonstrate that a classifier trained on a relatively balanced dataset in terms of age and sex already shows biased results on the held-out test set. In terms of \cite{corbett2018measure}, we apply the principle of \textit{classification parity}, meaning that we aim at making the predictive performance (AUC in our case) equal across subgroups defined by so-called protected features (in our case sex and age). As is also explained by \cite{corbett2018measure}, this mitigation strategy is not necessarily synonymous with fair machine learning, since equitable and fair decisions are very much context-dependent. Finally, we provide some guidelines for evaluating algorithms concerning this important topic. 

\subsection{Related work}
One form of dataset bias refers to a distribution shift between datasets, such that models trained on one dataset, show a drop in performance on the other. This idea has been studied in computer vision~\cite{torralba2011unbiased,khosla2012undoing}. In medical imaging, such drops in performance can be experienced in datasets collected at different centers~\cite{wachinger2018detect,ashraf2018learning,pooch2019can}. Such differences are often addressed with transfer learning~\cite{cheplygina2019not} techniques, which either align the data distributions or learn dataset-independent representations.

A more specific case of dataset bias is when the bias is based on the demographics of the training subjects including differences in ages, sexes, diets, habits, genetics and so on. Collected data often inadvertently encode human preferences.  
As an example, it has been demonstrated that face recognition algorithms can discriminate based on e.g. skin color and perform poorly on under-represented groups~\cite{buolamwini2018gender}. In medical imaging, similar factors might influence the data, thus have an impact on the incidence of disease too, as shown in some studies. For instance, \cite{COLE2017115,AgeFromRetina2019} show that signs of brain aging as a biomarker of aging can be predicted from brain and retinal images; or the work by~\cite{Cole1349} demonstrates the relation between the human immunodeficiency virus (HIV) and the aging process of the brain. A paper published after the first version of ours, shows gender bias in classification of lung diseases from chest X-ray images~\cite{larrazabal2020gender}. Thus it is essential to include or at least take into account the demographics in the data analysis. 

Various algorithms to mitigate this type of bias have been proposed. The first set of approaches focuses on preventing this bias in the first place i.e., creating a balanced set in the data preparation step~\cite{InterventionalFairness2019}. However, this is not always an option especially for medical data which is rare and where new acquisitions are often costly. 
Therefore, recent studies have focused mainly on learning representations that are not only predictive of the actual outputs but also invariant to the extraneous factors~\cite{pmlr-FairRepresentation,wang2019iccv,Roy2019CVPR,BiasResilient2020}.
In most cases, by including the additional available demographic information during training, their predictive power is mitigated by an adversarial loss and the features become invariant to them.

Due to the rise of machine learning diagnostic applications in the medical image analysis domain, we conduct a survey of the published techniques in MICCAI 2018~\cite{frangi2018medical} to investigate the inclusion of demographics in addition to the medical images. Our results show that even though the demographics might impact the outcome of the models, it is not a widely discussed topic in medical imaging. Not only most of the datasets do not include the demographic information, but also the proposed techniques rarely propose to correct for potential biases in their models. Additionally, we use a relatively balanced dataset of skin lesions~\cite{codella2017skin} and highlight the importance of correction for age and sex biases in this dataset. The closest study to this analysis is \cite{kinyanjui2019estimating}, where they show that skin lesion datasets over-represent lighter skin, but do not find large differences in performance for different skin types.  

\section{Methods}

\subsection{Paper analysis}

We screened the MICCAI 2018 proceedings~\cite{frangi2018medical} for papers on diagnosis using macroscopic images. We, therefore, focused on the chapters  ``Machine Learning in Medical Imaging'', ``Optical and Histology Applications'', ``Cardiac, Chest and Abdominal Applications'' and ``Neuroimaging and Brain Segmentation Methods: Neuroimaging''. Papers were included if they focused on the diagnosis or detection of abnormalities. For each selected paper, one of the authors quantified the following: number of public or private datasets used, number of subjects, whether demographic information was given, and whether demographics were discussed. 

\subsection{Classifier analysis}

To understand potential differences in the performance of a classifier for different demographic groups, we set up a baseline binary classification experiment. We used the ISIC 2017 skin lesion dataset~\cite{codella2017skin} for the diagnosis of melanoma skin cancer since the age and sex were available for over 75\% of the subjects. We included only the subjects for which both variables were available in our analysis. Age was provided to the nearest 5 years. To create large enough subgroups for evaluation, we split the subjects by calculating the median age in the training set (equal to 60) and using that as a threshold. The numbers of subjects in each group are provided in Table~\ref{tab:isic}. 

\begin{table}[]
    \centering
    \setlength\tabcolsep{0.25cm}
    \caption{Demographics of the used datasets.}
    \begin{tabular}{l r r r r r r}
    \hline
    ISIC subset & Total & Included & Male & Female & $<60$ & $\geq60$ \\
    \hline 
    Train          & 2000 &  1744 & 886 & 858 & 1087 & 657 \\
    Validation     & 150  & 149   & 90 & 59 & 87 & 62 \\
    Test           & 600 & 553 & 283 & 270 & 302 & 251 \\
    \hline
    \end{tabular}
    \label{tab:isic}
\end{table}


\textbf{Baseline network}. We trained an Inception-v4~\cite{szegedy2017inception} network as our baseline model using the training procedure from \cite{perez2018data}, which has outperformed the top result (0.874) from the ISIC 2017 challenge. The network uses data augmentation based on adjusting the color (saturation, contrast, brightness, hue) and geometry (affine transformations, flips, random crops) of the image. It is initialized with ImageNet weights, and then further trained on randomly augmented training images resized to 299 $\times$ 299. Training is then done with stochastic gradient descent with a momentum factor of 0.9, batch size of 40, and learning rate of 1e-3 which is reduced to 1e-4 after the 10th epoch. Early stopping is used if the validation area under the curve (AUC) does not improve after 8 epochs. At test time, an image is randomly augmented 32 times, and the predictions are averaged. All parameters are used as defined by \cite{perez2018data} and not specifically optimized for the subset of data that we used. We evaluated the classifiers with AUC for the following groups: all subjects, male, female, young ($<60$) and old ($\geq60$). 

\textbf{Bias-aware network}. To evaluate whether the learned representation has any relations to the available demographics, we use the method proposed by~\cite{BiasResilient2020}. Thus we employ an ensemble network with a shared feature encoder (the same as the baseline model) and two classifier heads.
One classifier is in charge of  classifying the skin cancer and it consists of a fully connected layer followed by average pooling and softmax layers (similar to the baseline model). The other head is supposed to predict the confounding parameter and it consists of a fully connected layer followed by an average pooling layer. Parameters of the encoder, cancer classifier and bias predictor are denoted by $\theta_e$, $\theta_c$, $\theta_{bp}$ respectively. Three losses are used for training the network. For training the skin cancer classifier head and encoder a cross-entropy loss ($L_c$) is used. While for optimizing the bias predictor head, a bias prediction loss ($L_{bp}$) is defined as the negative-squared Pearson correlation coefficient ($-Corr^2$). By minimizing $-Corr^2$, the correlation between the predicted and true confounding parameter should increase. Since sex is a binary parameter, in some experiments we define $L_{bp}$ as a binary cross-entropy loss ($BCE$). The third loss is defined as $L_{br}=-\lambda L_{bp}$ and is used to optimize the encoder adversarially to reduce the predictive power of the encoded features for the confounding parameter. $\lambda$ determines how much the encoder is penalized for leading to correct predictions of the target demographic parameter.

The ensemble network is trained iteratively with three main steps: 
\begin{enumerate*}[label={(\alph*)}]
\item updating $\theta_e$ and $\theta_c$ based on the $L_c$ loss;
\item updating only the $\theta_{bp}$ parameters based on $L_{bp}$ loss;
\item and finally updating $\theta_e$ adversarially based on $L_{br}$ loss
\end{enumerate*}. Note that the encoder weights are not updated in the second step, and the bias predictor weights are not updated in the third step. 
The updates are done one-by-one iteratively. The learning rates and optimizers of the three update steps are the same as the baseline model.  It is worth mentioning that for the steps involving the bias prediction, we only use the control data to make sure that the confounding parameters are reliably estimated from healthy subjects. Multiple experiments are performed to see whether it is possible to weaken the potential relationship between the encoded features from images and the confounding parameters, in our case age or sex. 


\section{Results}

\subsection{Paper analysis}

\begin{figure}
    \centering
    \hspace{-0.4cm}
    \includegraphics[width=0.51\textwidth]{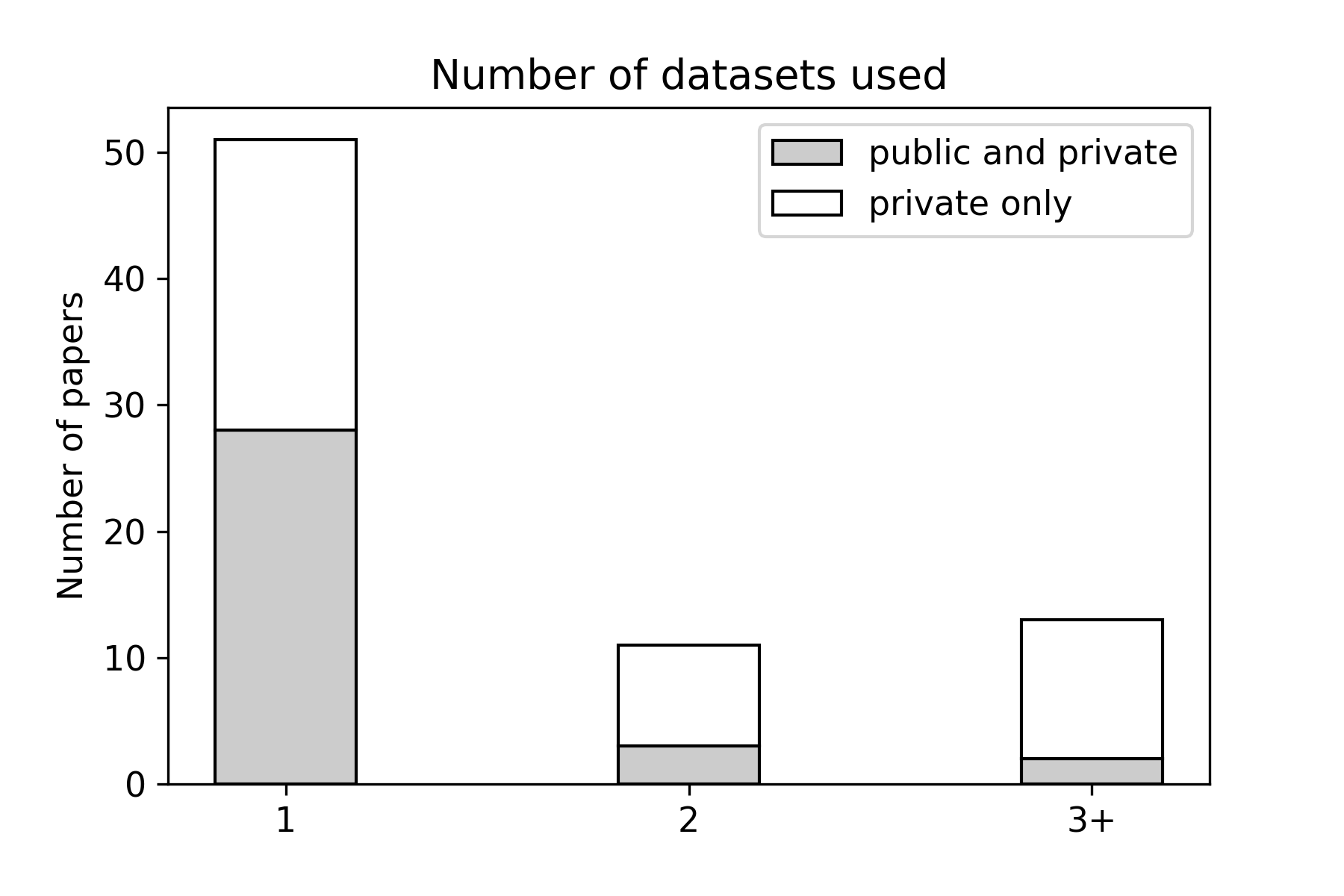}
    \hspace{-0.3cm}
    \includegraphics[width=0.51\textwidth]{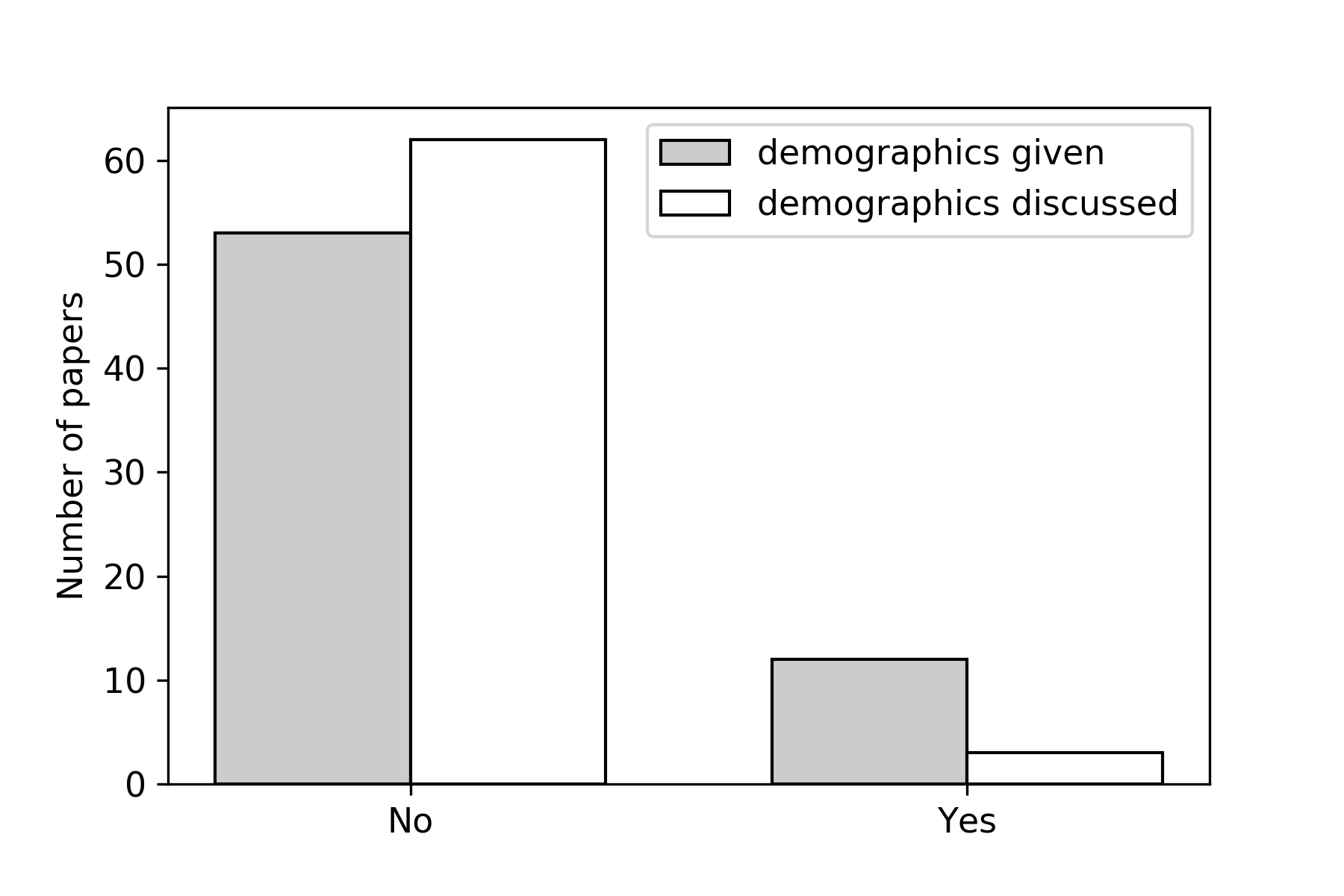}
    \caption{Number of papers using a certain number of private/public datasets (left), and including demographic information (right). }
    \label{fig:papers_datasets}
\end{figure}

A total of 65 papers fit our inclusion criteria. Several statistics of the datasets used, and the inclusion of demographic information by the papers are shown in Fig.~\ref{fig:papers_datasets}. In total there were 52 papers using 1 dataset, 11 papers using 2 datasets, and 2 papers using 3 or more datasets. Nearly half (32 papers) did not use any public datasets. The sizes of the datasets varied between 10 subjects and 112K subjects, with 217 subjects as the median size.  

In this set of 65 papers, 12 papers described at least age or sex. Notably, 10 of these were neuroimaging papers. Of the 12 papers, only 3 also evaluate or discuss their results with respect to the demographics. \cite{orlando2018towards} test whether their glaucoma risk index differs significantly between the healthy and patient groups, while also checking whether these groups have statistically different age and sex distributions. \cite{gill2018deep} stratify their results of detecting brain malformations by age group (children vs adults). Finally \cite{hett2018graph} corrects their Alzheimer's score estimation for brain images, with a factor based on linear regression of cognitively normal subjects.

\subsection{Classifier analysis}

\begin{table}[t]
\centering 
\caption{An overview of the AUCs obtained in each experiment. The most balanced performances after correction for the bias are bolded.}\label{tab1}
\setlength\tabcolsep{0.1cm}
\begin{tabular}{ccccccccc}
\hline 
\textbf{Experiment} & \textbf{Confounder} & \textbf{$\lambda$} & \textbf{$L_{bp}$}  & \textbf{All} & \textbf{Young} & \textbf{Old} & \textbf{Male} & \textbf{Female}  \\
\hline 
1. Baseline & N/A & N/A & N/A & 0.83 &	0.76 &	0.84 &	0.76 &	0.90  \\
\hline 
2. Ensemble & age & 0  & $-Corr^2$ & 0.83 & 	0.78 &	0.83	& 0.77 & 	0.90  \\
3. Ensemble & age & 5 &  $-Corr^2$ &  0.80 &	\textbf{0.77} &	\textbf{0.80} &	0.73 &	0.90  \\
\hline 
4. Ensemble & sex & 0 & $-Corr^2$ & 0.83 &	0.77 &	0.84 &	0.76 &	0.91  \\
5. Ensemble & sex &  5 &$-Corr^2$ & 0.72 &	0.64 & 	0.75 & 	\textbf{0.73} &	\textbf{0.78} \\
\hline 
6. Ensemble & sex & 0 &   $BCE$ & 0.84 &	0.77 &	0.85 &	0.78 &	0.90 \\
7. Ensemble & sex & 0.5 &  $BCE$  & 0.83 &	0.78 &	0.84 &	0.77 &	0.91  \\
\hline
\end{tabular}
\label{tab:results}
\end{table}

\begin{figure}[t]
    \centering
    \hspace{-0.5cm}
    \includegraphics[width=0.53\textwidth]{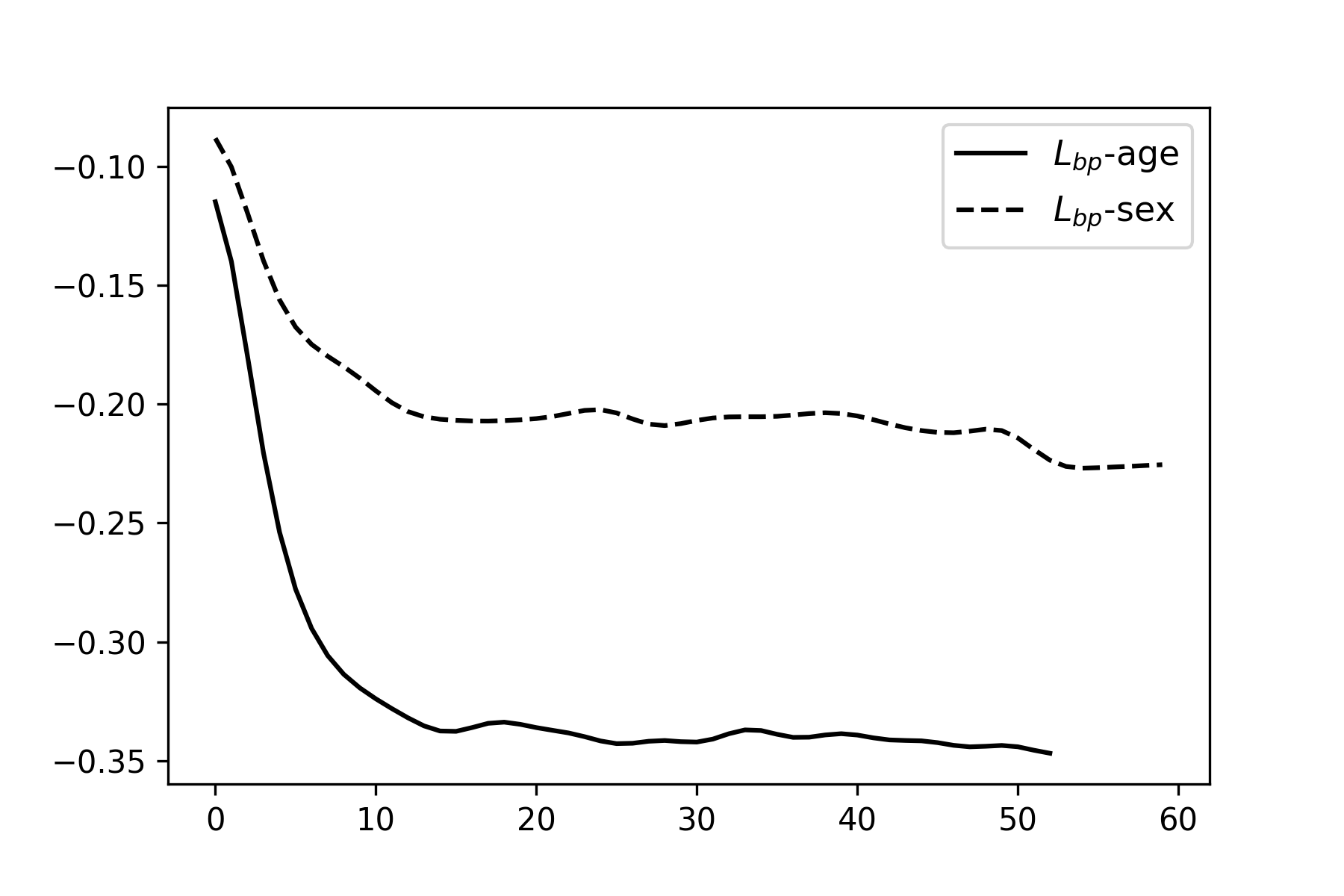} 
    \hspace{-0.5cm}
    \includegraphics[width=0.53\textwidth]{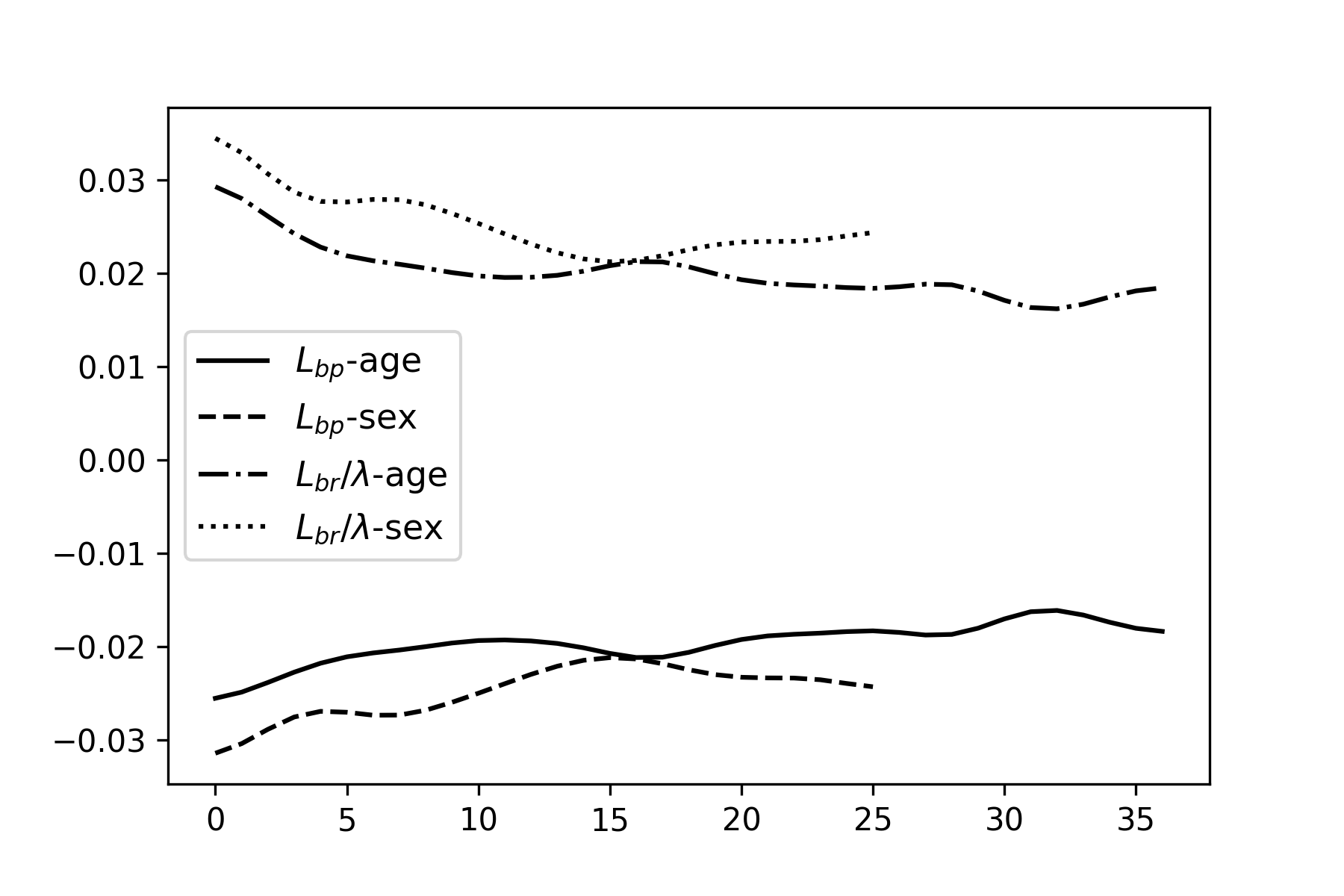}
    \hspace{-0.5cm}
    \caption{Left: The $L_{bp}$ loss of the ensemble network with $\lambda=0$ (experiments 2 and 4 in Table~\ref{tab:results}); right: The $L_{bp}$ and $L_{br}/\lambda$ losses when $\lambda > 0$ (experiments 3 and 5). }
    \label{fig:lbp_loss}
\end{figure}

The AUC performances on the test set across different subgroups and all the subjects are shown in
Table~\ref{tab:results}. For each experiment, we have specified the $L_{bp}$, demographic,  and $\lambda$ parameters used during training. The baseline model achieves an overall AUC of  0.83 that is slightly lower than the 
AUC of 0.88 reported in \cite{perez2018data} because we only use the subset of subjects with known demographics and we do only half of the test time augmentations used by \cite{perez2018data}. Moreover, the ensemble networks achieve the same performances as the baseline model when $\lambda$ is set to 0 (experiments 2, 4 and 6) because there is no back-propagation from the bias predictor head to the encoder i.e., $L_{br}=0$. In these experiments, both subgroup splits show large differences between them: depending on age, the AUC varies between 0.76 and 0.85 (9\%), and depending on sex, between 0.76 and 0.91 (15\%). The obtained $L_{bp}$ values for experiment 2 and 4 ($\lambda=0$) are also shown in Fig.~\ref{fig:lbp_loss} (left). As seen in this figure, there exist some correlations between the predicted  and true  confounding parameter when the encoder is only optimized for classifying skin cancer. This correlation is stronger for age than sex.

When we use the ensemble network to predict the age of the control subjects in an adversarial setting (experiment 3), we see that the differences between the performances of young and old subgroups decrease (only 3\%), while that is not the case for male/female subgroups. Similarly, when the sex is used as the confounding parameter (experiment 5), the AUC of male/female subgroups get closer (5\% difference), while the difference between young/old subgroups remains large (11\%).  The values of $L_{bp}$ and $L_{br}/\lambda$ for experiments 3 and 5 are visualized in Fig.~\ref{fig:lbp_loss} (right). Since the training is a min-max optimization problem,  compared to the left figure, the correlation does not increase during training i.e., $L_{bp}$ does not decrease. Additionally, the training stops much earlier resulting in a drop in the overall AUC of the skin cancer classifier.

Since sex is a binary parameter unlike age, the $BCE$ loss is used in experiments 6 and 7. As depicted by results, the $BCE$ loss is not as effective as the $-Corr^2$ and the AUCs are almost the same as the baseline model. Note that $\lambda$ is determined heuristically based on the ratio between $L_c$ and $L_{bp}$ loss in order to have an effective penalty in updating the encoder weights. 

\section{Discussion and conclusions}


Our paper analysis showed that demographics are rarely discussed and used in diagnostic algorithms. 
This is surprising, given the importance of demographics in diagnosis. For example, men and women have different distributions of melanoma subtypes~\cite{beddingfield2003melanoma}, which can affect the final diagnosis.

Our classifier analysis results showed large differences in performance between male and female subjects, and between different groups of age for the baseline model. The male/female difference is somewhat surprising, given that the training data was relatively balanced. This suggests that these factors might influence how difficult a skin lesion is to diagnose. 

Additionally, we demonstrated the possibility to correct for the potential bias in predictions to some extent by using an adversarial training setting. The same method was used in \cite{BiasResilient2020} to investigate whether diagnosis of HIV based on brain MRIs is dependent on subjects' age instead of true HIV markers or not.  Their results suggest that predictions from the baseline may be biased, whereas the bias-aware network results in a space with no apparent bias to age.

Our results indicate that age, sex and possibly other characteristics might bias the results differently. There might be some correlations between different confounders, or a case of Simpson's paradox~\cite{wagner1982simpson}. Moreover, there might be additional unknown factors (for instance the skin color or the hairs on the skin) that need to be identified and treated appropriately. In general, correction is more effective, when all confounding parameters are known and used simultaneously along with training for the main target task. Additionally, we treated the age as a continuous parameter, but the evaluation was done for two subgroups (young and old). The fairness of this evaluation strategy needs to be investigated in future works.


 A possible way to address the bias problem would be to standardize what information about the data or model needs to be included in a research paper. This could be inspired by datasheets for datasets \cite{datasheets2018}, which describes the dataset design and collection procedure, and relatedly model cards~\cite{mitchell2019model}, which describe in detail the choices made to train and optimize the model. Enforcing such standards would require large-scale collaboration between journals and conferences, but researchers could already include such information to increase awareness in the community as a whole. Although this type of measure does not remove bias, it can show that a bias potentially might exist. The exact sources of this bias could then be quantified, for example following the framework proposed by~\cite{suresh2019framework}.  

A recent interesting approach is the development of tools for assessing and mitigating the amount of bias and fairness in machine learning models and datasets. For example, Aequitas \cite{Saleiro2018-jl} is an open source toolkit developed at Center for Data Science and Public Policy University of Chicago in order to enable users to test models for several bias and fairness metrics in relation to multiple population sub-groups. Another example is the AI Fairness 360 (AIF360) \cite{Bellamy2019-de} open source toolkit for checking unwanted bias and moreover mitigating it. This toolkit developed at IBM focuses on industrial usability and software engineering. These tools can be helpful for data scientists, researchers, policy makers and software engineers.

Another important direction is building bias-aware algorithms and removing the bias in the final predictions. Even though that might be at the expense of overall model accuracy~\cite{Zemel2013,Kamishima2011}. Thus dataset and model interventions are both necessary~\cite{towardsFairness2020}. Once an algorithm is designed to be sensitive to bias, we need to evaluate whether it is successful at this. Therefore, we need ways to quantify what performance gap is evidence of bias or not.

In conclusion, we highlighted the importance of bias in medical datasets and diagnostic algorithms, since ignoring it could affect the generalization across different demographic subgroups. 
We believe that this is an important point of attention for researchers working in medical image analysis community.

\bibliographystyle{splncs04}
\bibliography{refs_veronika,refs_public}

\begin{thebibliography}{10}
\providecommand{\url}[1]{\texttt{#1}}
\providecommand{\urlprefix}{URL }
\providecommand{\doi}[1]{https://doi.org/#1}

\bibitem{BiasResilient2020}
Adeli, E., Zhao, Q., Pfefferbaum, A., Sullivan, E.V., Fei-Fei, L., Niebles,
  J.C., Pohl, K.M.: Representation learning with statistical independence to
  mitigate bias (2019)

\bibitem{ashraf2018learning}
Ashraf, A., Khan, S., Bhagwat, N., Chakravarty, M., Taati, B.: Learning to
  unlearn: Building immunity to dataset bias in medical imaging studies. arXiv
  preprint arXiv:1812.01716  (2018)

\bibitem{beddingfield2003melanoma}
Beddingfield~3rd, F.: The melanoma epidemic: res ipsa loquitur. The Oncologist
  \textbf{8}(5), ~459 (2003)

\bibitem{Bellamy2019-de}
Bellamy, R.K.E., Dey, K., Hind, M., Hoffman, S.C., Houde, S., Kannan, K.,
  Lohia, P., Martino, J., Mehta, S., Mojsilovi{\'c}, A., Nagar, S., Ramamurthy,
  K.N., Richards, J., Saha, D., Sattigeri, P., Singh, M., Varshney, K.R.,
  Zhang, Y.: {AI} fairness 360: An extensible toolkit for detecting and
  mitigating algorithmic bias. IBM J. Res. Dev.  \textbf{63}(4/5),  4:1--4:15
  (Jul 2019)

\bibitem{buolamwini2018gender}
Buolamwini, J., Gebru, T.: Gender shades: Intersectional accuracy disparities
  in commercial gender classification. In: Conference on Fairness,
  Accountability and Transparency. pp. 77--91 (2018)

\bibitem{cheplygina2019not}
Cheplygina, V., de~Bruijne, M., Pluim, J.P.: Not-so-supervised: a survey of
  semi-supervised, multi-instance, and transfer learning in medical image
  analysis. Medical image analysis  \textbf{54},  280--296 (2019)

\bibitem{codella2017skin}
Codella, N.C., Gutman, D., Celebi, M.E., Helba, B., Marchetti, M.A., Dusza,
  S.W., Kalloo, A., Liopyris, K., Mishra, N., Kittler, H., et~al.: Skin lesion
  analysis toward melanoma detection: A challenge at the 2017 {International
  Symposium on Biomedical Imaging (ISBI)}, hosted by the {International Skin
  Imaging Collaboration (ISIC)}. arXiv preprint arXiv:1710.05006  (2017)

\bibitem{COLE2017115}
Cole, J.H., Poudel, R.P., Tsagkrasoulis, D., Caan, M.W., Steves, C., Spector,
  T.D., Montana, G.: Predicting brain age with deep learning from raw imaging
  data results in a reliable and heritable biomarker. NeuroImage  \textbf{163},
   115 -- 124 (2017)

\bibitem{Cole1349}
Cole, J.H., Underwood, J., et~al.: Increased brain-predicted aging in treated
  {HIV} disease. Neurology  \textbf{88}(14),  1349--1357 (2017)

\bibitem{corbett2018measure}
Corbett-Davies, S., Goel, S.: The measure and mismeasure of fairness: a
  critical review of fair machine learning. arXiv preprint arXiv:1808.00023
  (2018)

\bibitem{pmlr-FairRepresentation}
Creager, E., Madras, D., Jacobsen, J.H., Weis, M., Swersky, K., Pitassi, T.,
  Zemel, R.: Flexibly fair representation learning by disentanglement. In:
  Chaudhuri, K., Salakhutdinov, R. (eds.) International Conference on Machine
  Learning. Proceedings of Machine Learning Research, vol.~97, pp. 1436--1445.
  PMLR, Long Beach, California, USA (09--15 Jun 2019)

\bibitem{frangi2018medical}
Frangi, A.F., Schnabel, J.A., Davatzikos, C., Alberola-L{\'o}pez, C.,
  Fichtinger, G.: Medical Image Computing and Computer Assisted
  Intervention--{MICCAI} 2018: 21st International Conference, Granada, Spain,
  September 16-20, 2018, Proceedings, vol. 11073. Springer (2018)

\bibitem{datasheets2018}
Gebru, T., Morgenstern, J., Vecchione, B., Vaughan, J.W., Wallach, H.M., III,
  H.D., Crawford, K.: Datasheets for datasets. CoRR  \textbf{abs/1803.09010}
  (2018)

\bibitem{gill2018deep}
Gill, R.S., Hong, S.J., Fadaie, F., Caldairou, B., Bernhardt, B.C., Barba, C.,
  Brandt, A., Coelho, V.C., d'Incerti, L., Lenge, M., et~al.: Deep
  convolutional networks for automated detection of epileptogenic brain
  malformations. In: Medical Image Computing and Computer-Assisted
  Intervention. pp. 490--497. Springer (2018)

\bibitem{greenspan2016guest}
Greenspan, H., Van~Ginneken, B., Summers, R.M.: Guest editorial deep learning
  in medical imaging: Overview and future promise of an exciting new technique.
  IEEE Transactions on Medical Imaging  \textbf{35}(5),  1153--1159 (2016)

\bibitem{hett2018graph}
Hett, K., Ta, V.T., Manj{\'o}n, J.V., Coup{\'e}, P., Initiative, A.D.N.,
  et~al.: Graph of brain structures grading for early detection of alzheimer's
  disease. In: Medical Image Computing and Computer-Assisted Intervention. pp.
  429--436. Springer (2018)

\bibitem{Kamishima2011}
{Kamishima}, T., {Akaho}, S., {Sakuma}, J.: Fairness-aware learning through
  regularization approach. In: 2011 IEEE 11th International Conference on Data
  Mining Workshops. pp. 643--650 (2011)

\bibitem{khosla2012undoing}
Khosla, A., Zhou, T., Malisiewicz, T., Efros, A.A., Torralba, A.: Undoing the
  damage of dataset bias. In: European Conference on Computer Vision. pp.
  158--171. Springer (2012)

\bibitem{kinyanjui2019estimating}
Kinyanjui, N.M., Odonga, T., Cintas, C., Codella, N.C., Panda, R., Sattigeri,
  P., Varshney, K.R.: Estimating skin tone and effects on classification
  performance in dermatology datasets. arXiv preprint arXiv:1910.13268  (2019)

\bibitem{larrazabal2020gender}
Larrazabal, A.J., Nieto, N., Peterson, V., Milone, D.H., Ferrante, E.: Gender
  imbalance in medical imaging datasets produces biased classifiers for
  computer-aided diagnosis. Proceedings of the National Academy of Sciences
  (2020)

\bibitem{AgeFromRetina2019}
Liu, C., Wang, W., Li, Z., Jiang, Y., Han, X., Ha, J., Meng, W., He, M.:
  Biological age estimated from retinal imaging: A novel biomarker of aging.
  In: Shen, D., Liu, T., Peters, T.M., Staib, L.H., Essert, C., Zhou, S., Yap,
  P.T., Khan, A. (eds.) Medical Image Computing and Computer Assisted
  Intervention. pp. 138--146. Springer International Publishing, Cham (2019)

\bibitem{mitchell2019model}
Mitchell, M., Wu, S., Zaldivar, A., Barnes, P., Vasserman, L., Hutchinson, B.,
  Spitzer, E., Raji, I.D., Gebru, T.: Model cards for model reporting. In:
  Fairness, Accountability, and Transparency (FAccT). pp. 220--229. ACM (2019)

\bibitem{orlando2018towards}
Orlando, J.I., Breda, J.B., Van~Keer, K., Blaschko, M.B., Blanco, P.J., Bulant,
  C.A.: Towards a glaucoma risk index based on simulated hemodynamics from
  fundus images. In: Medical Image Computing and Computer-Assisted
  Intervention. pp. 65--73. Springer (2018)

\bibitem{perez2018data}
Perez, F., Vasconcelos, C., Avila, S., Valle, E.: Data augmentation for skin
  lesion analysis. In: OR 2.0 Context-Aware Operating Theaters, Computer
  Assisted Robotic Endoscopy, Clinical Image-Based Procedures, and Skin Image
  Analysis, pp. 303--311. Springer (2018)

\bibitem{pooch2019can}
Pooch, E.H., Ballester, P.L., Barros, R.C.: Can we trust deep learning models
  diagnosis? the impact of domain shift in chest radiograph classification.
  arXiv preprint arXiv:1909.01940  (2019)

\bibitem{Roy2019CVPR}
{Roy}, P.C., {Boddeti}, V.N.: Mitigating information leakage in image
  representations: A maximum entropy approach. In: Computer Vision and Pattern
  Recognition (CVPR). pp. 2581--2589 (June 2019)

\bibitem{Saleiro2018-jl}
Saleiro, P., Kuester, B., Hinkson, L., London, J., Stevens, A., Anisfeld, A.,
  Rodolfa, K.T., Ghani, R.: Aequitas: A bias and fairness audit toolkit. arXiv
  preprint arXiv:1811.05577  (Nov 2018)

\bibitem{InterventionalFairness2019}
Salimi, B., Rodriguez, L., Howe, B., Suciu, D.: Interventional fairness: Causal
  database repair for algorithmic fairness. In: International Conference on
  Management of Data. pp. 793---810. Association for Computing Machinery (2019)

\bibitem{suresh2019framework}
Suresh, H., Guttag, J.V.: A framework for understanding unintended consequences
  of machine learning. arXiv preprint arXiv:1901.10002  (2019)

\bibitem{szegedy2017inception}
Szegedy, C., Ioffe, S., Vanhoucke, V., Alemi, A.A.: Inception-v4,
  inception-resnet and the impact of residual connections on learning. In: AAAI
  Conference on Artificial Intelligence (2017)

\bibitem{torralba2011unbiased}
Torralba, A., Efros, A.A.: Unbiased look at dataset bias. In: Computer Vision
  and Pattern Recognition (CVPR). pp. 1521--1528 (2011)

\bibitem{wachinger2018detect}
Wachinger, C., Becker, B.G., Rieckmann, A.: Detect, quantify, and incorporate
  dataset bias: A neuroimaging analysis on 12,207 individuals. arXiv preprint
  arXiv:1804.10764  (2018)

\bibitem{wagner1982simpson}
Wagner, C.H.: Simpson's paradox in real life. The American Statistician
  \textbf{36}(1),  46--48 (1982)

\bibitem{wang2019iccv}
Wang, T., Zhao, J., Yatskar, M., Chang, K.W., Ordonez, V.: Balanced datasets
  are not enough: Estimating and mitigating gender bias in deep image
  representations. In: International Conference on Computer Vision (2019)

\bibitem{towardsFairness2020}
Yang, K., Qinami, K., Fei-Fei, L., Deng, J., Russakovsky, O.: Towards fairer
  datasets: Filtering and balancing the distribution of the people subtree in
  the imagenet hierarchy. In: Proceedings of the 2020 Conference on Fairness,
  Accountability, and Transparency. pp. 547--558. FAT* '20, Association for
  Computing Machinery, New York, NY, USA (2020)

\bibitem{Zemel2013}
Zemel, R., Wu, Y., Swersky, K., Pitassi, T., Dwork, C.: Learning fair
  representations. In: Dasgupta, S., McAllester, D. (eds.) Proceedings of the
  30th International Conference on Machine Learning. Proceedings of Machine
  Learning Research, vol. 28(3), pp. 325--333. PMLR, Atlanta, Georgia, USA
  (17--19 Jun 2013)

\end{thebibliography}
\end{document}